\documentclass{article}
\usepackage{amssymb}

\usepackage[preprint]{corl_2025} 

\usepackage{graphics}
\usepackage{graphicx}
\usepackage{epsfig}
\usepackage{mathptmx}
\usepackage{times}
\usepackage{amsmath}
\usepackage{amssymb}
\usepackage{rotating}
\usepackage{array}
\usepackage{multirow}
\usepackage{makecell}
\usepackage{multicol}
\usepackage{booktabs}
\usepackage{makecell}
\usepackage{colortbl}
\usepackage{hyperref}
\usepackage{algorithm}
\usepackage{algorithmic}
\usepackage{bbding}
\usepackage{enumitem}

\title{Bootstrapping Imitation Learning for Long-horizon Manipulation via Hierarchical Data Collection Space}

\author{
  Jinrong Yang$^1$$^2$$^*$, Kexun Chen$^3$$^*$, Zhuoling Li$^4$, Shengkai Wu$^1$, Yong Zhao$^1$, Liangliang Ren$^1$\\ \textbf{Wenqiu Luo$^1$, Chaohui Shang$^1$, Meiyu Zhi$^1$, Linfeng Gao$^1$, Mingshan Sun$^1$, Hui Cheng$^2$$^\dagger$}\\
  $^1$CVTE \qquad $^2$SYSU \qquad  $^3$SWJTU \qquad HKU\\
  $^*$Equal contribution \quad\quad $^\dagger$Corresponding author
}

\begin{document}
\maketitle

\begin{abstract}
Imitation learning (IL) with human demonstrations is a promising method for robotic manipulation tasks. While minimal demonstrations enable robotic action execution, achieving high success rates and generalization requires high cost, \emph{e.g.}, continuously adding data or incrementally conducting human-in-loop processes with complex hardware/software systems. In this paper, we rethink the state/action space of the data collection pipeline as well as the underlying factors responsible for the prediction of non-robust actions.
To this end, we introduce a \textbf{\underline{H}ierarchical \underline{D}ata Collection \underline{Space} (HD-Space)} for robotic imitation learning, a simple data collection scheme, endowing the model to train with proactive and high-quality data. 
Specifically, We segment the fine manipulation task into multiple key atomic tasks from a high-level perspective and design atomic state/action spaces for human demonstrations, aiming to generate robust IL data.
We conduct empirical evaluations across two simulated and five real-world long-horizon manipulation tasks and demonstrate that IL policy training with HD-Space-based data can achieve significantly enhanced policy performance.
HD-Space allows the use of a small amount of demonstration data to train a more powerful policy, particularly for long-horizon manipulation tasks. 
We aim for HD-Space to offer insights into optimizing data quality and guiding data scaling. We provide supplementary materials and
videos on our \href{https://hd-space-robotics.github.io}{project page}.
\end{abstract}


\section{Introduction}
Imitation Learning (IL) is a promising way to learn elaborate and long-horizon manipulation tasks~\citep{pomerleau1988alvinn,hussein2017imitation,jang2022bc,mandlekar2021matters,zhao2023learning,chi2023diffusion,ze20243d,goyal2023rvt}. Experts leverage teleoperation interfaces~\citep{zhao2023learning, umi,fu2024mobile,fu2024humanplus,cheng2024open} to gather successful demonstrations for policy learning. However, this approach typically demands a large dataset to ensure the policy generalizes effectively to uncertainties like object spatial pose~\citep{lin2024data, lillicrap2015continuous,tan2024manibox}. Ideally, a comprehensive set of demonstrations covering diverse states (\emph{e.g.}, image views, and robotic proprioception information) is needed for robust generalization. Yet, collecting such a large-scale dataset is costly, limiting the practical application of imitation learning.

To solve this issue, recent approaches attempt to address it via Human-in-the-Loop (HIL)~\citep{mandlekar2020human,luo2024precise,hiranaka2023primitive} or Real2Sim2Real~\citep{li2024robogsim,katara2024gen2sim,wang2023real2sim2real1,wu2024rl,han2025re,wang2023real2sim2real,zhang2023efficient} pipelines. HIL collects out-of-distribution (OOD) data by specifically observing and then correcting the policy's wrong trajectories for retraining. While this approach can avoid collecting redundant data, it has two drawbacks: a) Experts have to passively wait for errors to occur before they can trigger data correction, which affects the efficiency of the data correction process; b) Real-time correction data imposes high requirements on the construction of the hardware system. Unlike the HIL, the latter is an active data collection method with no upper limit on the amount of data. However, it requires precise object modeling and correspondence for real-world scenes, and there is still the sim-to-real gap issue in terms of scene authenticity and the physical dynamic modeling of objects.

\begin{figure*}[t]
    \centering
    \includegraphics[width=0.96\linewidth]{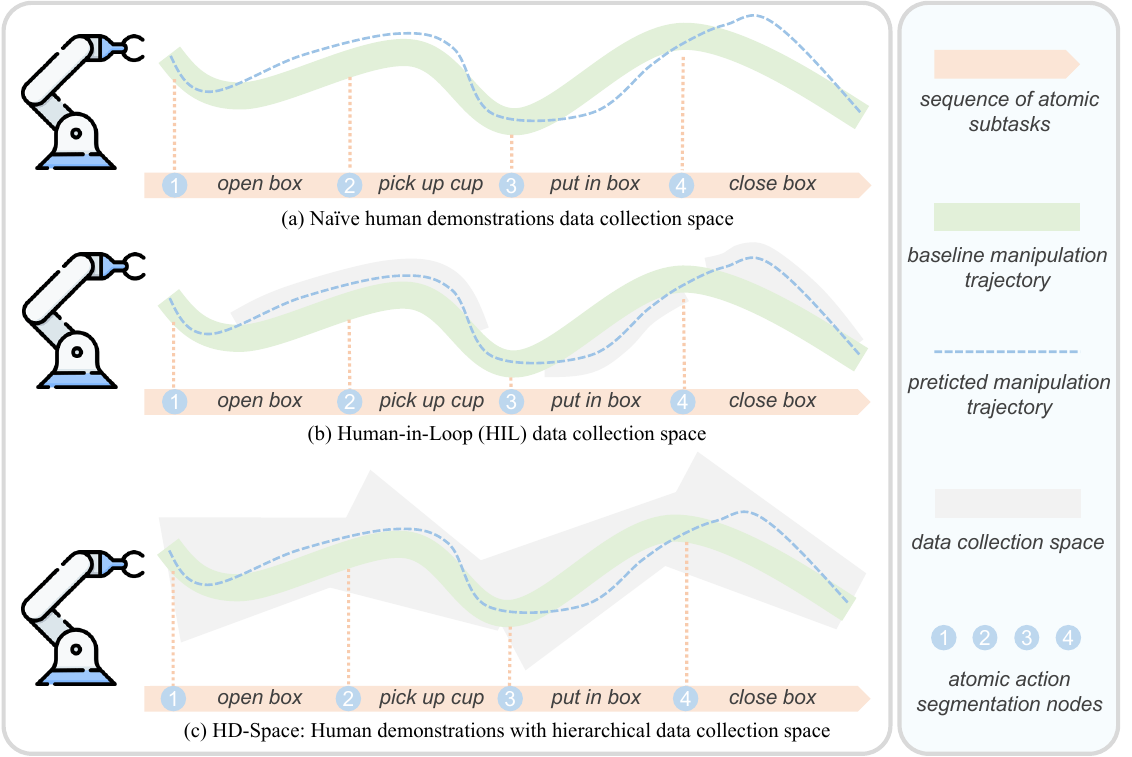}
    \vspace{-0.1cm}
    \caption{\textbf{Conceptual comparison} of human demonstration data spaces for imitation learning. For simplification, we use a 2D manipulation state/action space for elaboration. 
    (a) The naive data collection method records the entire manipulation trajectory, recording a single distribution of trajectories in the second to fourth stages. Therefore, the model is prone to incorrectly predicted trajectories in the subtasks of stages 2 to 4.
    (b) The HIL data collection method first deploys a baseline model for inference and then cooperates with the hardware system to correct the incorrect trajectories. The theoretical data collection space is the dynamic space where the model is prone to incorrect trajectories. Although the data collection becomes smaller, it still requires a complex hardware system and extra time to wait for the triggering of incorrect trajectories.
    (c) Our proposed HD-Space segments overall data collection space into multiple overlapped atomic spaces and then collects the manipulation data in each atomic space uniformly. It not only reduces the data collection space but also can efficiently traverse the data that are prone to errors in predicting trajectories after the subtasks of the first stage. Moreover, HD-Space can also provide a better baseline model for HIL.
 }\label{fig:overall}
\vspace{-1cm}
\end{figure*}

In this paper, we introduce a simple \textbf{\underline{H}ierarchical \underline{D}ata Collection \underline{Space} (HD-Space)} designed to enhance the robustness and efficiency of data collection for IL.
As illustrated in Fig.~\ref{fig:overall} (c), it initially segments tdivides complex manipulation tasks into atomic subtasks (\emph{e.g.}, picking, placing, opening, and closing, \emph{etc.}) from a macro perspective.
Starting with these atomic tasks within each state/action space, it systematically explores the initial end pose of the robotic arm, thereby accomplishing both the atomic task and the data collection procedure.
Importantly, the overlapping boundaries between the start and end of each atomic state/action space ensure the smooth execution of extended manipulation sequences.
HD-Space's refined task segmentation facilitates the acquisition of more reliable and concise datasets. 
Compared with conventional and HIL data collection methods depicted in Fig.~\ref{fig:overall} (a) and (c), HD-Space demonstrates two key advantages: robustness, achieved through active exploration of atomic manipulation spaces that minimize prediction errors as tasks progress; Efficiency, enabled by segmented state/action spaces and targeted data collection that reduces requirements for data volume, time and the learning price of demonstration.
The application of HD-Space for data collection offers valuable insights for advancing imitation learning in sophisticated manipulation scenarios.

The key contribution of this work is a robust and efficient data collection method for IL.
Through experiments across three real-world manipulation tasks, HD-Space proves effective in gathering smaller yet higher-quality datasets to train a better IL-based policy. We hope that HD-Space will offer fresh inspiration for optimizing data quality and guiding future data-scaling efforts.

\section{Related Work}
\label{sec:related_work}

\subsection{Imitation Learning for Manipulation}
\label{sec:imitation_learning}

Imitation learning (IL) for robotic manipulation has been extensively studied. This approach predominantly employs behavior cloning~\citep{pomerleau1988alvinn,schaal1999imitation,argall2009survey}, where a policy is directly learned from expert demonstrations by mapping states (\emph{e.g.}, images, point clouds, and robotic proprioceptions) to actions. However, this method often suffers from poor generalization on account of the trigger of compounding errors when the policy breaks away from states seen in the demonstration data~\citep{zhao2023learning,shi2023waypoint}. To address this problem, recent works aim to design stronger policy networks, like CNN~\citep{jang2022bc,james2022q,pari2021surprising}, Transformer~\citep{shafiullah2022behavior,zhu2023viola, zhao2023learning,lee2024interact}, Diffusion~\citep{chi2023diffusion,ze20243d,chen2025responsive,pearce2023imitating}, Vision-Language-Action (VLA) models~\citep{bharadhwaj2024roboagent,haldar2024baku,brohan2022rt,kim2024openvla,black2024pi_0}, \emph{etc}.
This paper instead focuses on designing the HD-Space to improve the quality of demonstration data for IL policy training, which is orthogonal to any new model architecture and even provides positive supplementation.

\subsection{Data Collection and Formulation for Imitation Learning}
\label{sec:data_collection}

Collecting high-quality data for IL training is the key to obtaining a better model. Some methods are dedicated to collecting data on a larger scale in real-world scene~\citep{lin2024data, tan2024manibox,bharadhwaj2024roboagent,o2024open,khazatsky2024droid,wu2024robomind} or in the reconstructed simulated scene~\citep{li2024robogsim,katara2024gen2sim,wang2023real2sim2real1,wu2024rl,han2025re,wang2023real2sim2real,zhang2023efficient}. It has the opportunity to enable the model to see more diverse image/point cloud observation and robot states, thereby enhancing the model's location generalization ability for a certain task. However, most of these data fix the position of the robotic arm and then collect the data based on the combination state of the random poses of the object, which is prone to cause trajectory compounding errors~\citep{shi2023waypoint,zhao2023learning,chi2023diffusion}, especially for long-horizon tasks. To this end, the HIL method is designed to correct the incorrect trajectory and re-train the model more robust~\citep{mandlekar2020human,luo2024precise,hiranaka2023primitive}. Although HIL is effective, it requires passively waiting for the trigger of error trajectories and conducting secondary training of the model, consuming data collection time and reducing the efficiency of model learning. Another solution to compound errors is to perform post-processing on the data, like resampling better waypoints among original action trajectories using different heuristics~\citep{james2022q,shi2023waypoint, shridhar2023perceiver, hsiao2006imitation, akgun2012keyframe,zeng2021transporter,shridhar2022cliport,morrison2018closing}. Recently, the work that is relatively relevant to us is ADC~\citep{huang2025adversarial}, which introduces an adversarial two-human-robot intervention for better data collection and improves the robustness of IL training.
In this paper, we start from the state/action space of fine or long-horizon manipulation tasks, and then segment atomic subtask and hierarchical space to explore the error-prone spaces of states and actions. This method streamlines the process, requiring solely one expert for human demonstration. Its hierarchical collection space approach facilitates data market acquisition in shorter units. This not only simplifies quality control for the expert but also makes it easier to communicate the essence of data collection to other operators. Moreover, the HD-Space is also orthogonal to and complementary to the above methods.

\section{HD-Space: Hierarchical Data Collection Space}
\label{sec:method}

In this section, we present the HD-Space data collection method. It is designed to enhance the robustness and efficiency of human demonstrations for robot manipulation imitation learning.

\textbf{Problem Formulation.} Given a set of human demonstrations $S$, each time $t$ includes visual observation $o_t$, robot state observation $p_t$ (\emph{e.g.}, joint angles or end-effector poses), and a temporal action sequence ($a_t$, $a_{t+1}$, $...$, $a_{t+N}$). The goal of IL is to learn a policy ${\pi}_{\theta}$ receives visual and robot state observations and predict a sequence of actions during training via the supervised loss:
\begin{equation}\label{eq:il_loss}
    \mathcal{L}(\theta) = -\mathbb{E}_{(o,p,a)\sim{S}_{O}}
    \left[
    \log {\pi}_{\theta} (a|o,p)
    \right].
\end{equation}

\textbf{Problem Analysis.} 
To learn a robust IL policy, high-quality demonstration data is critical~\citep{belkhale2023data,samadi2024good,lin2024data1,huang2025adversarial}. This paper focuses on improving the data collection process to acquire higher-quality data at a lower cost. We first analyze the conventional data collection approach. As shown in Fig.~\ref{fig:overall} (a), experts operate the robotic arm to complete the entire manipulation task and collect a piece of demonstration data, which only captures the optimal trajectory (the thick green area) under different object arrangements. Since the states/actions space outside the green area is not collected, the trained model is prone to predicting incorrect trajectories. Moreover, only traversing the states/actions of green areas leads to data redundancy and time consumption, making it easy to obtain a suboptimal model with the learning goal of Eq.~\ref{eq:il_loss}. To address this issue, the HIL method (Fig.~\ref{fig:overall} (b)) is a promising way to expand the data collection space. However, this process is still not efficient enough as it not only requires the establishment of a complex intervention system but also passively waits for the generation of incorrect trajectories, resulting in more time consumption.

\begin{figure*}[h]
    \centering
    \includegraphics[width=0.98\linewidth]{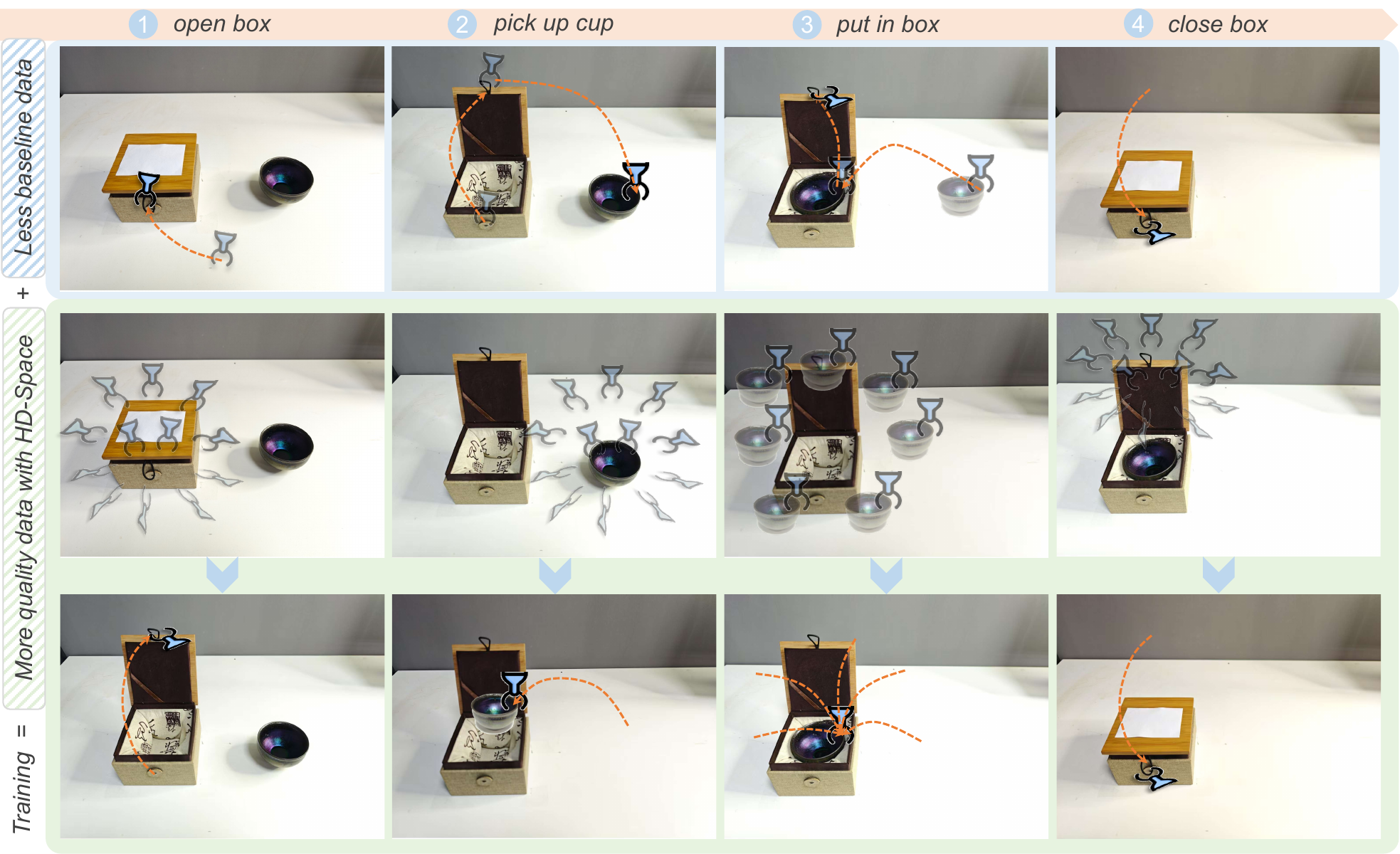}
    \vspace{-0.2cm}
    \caption{\textbf{Practice and comparison of naive and HD-Space} data collection methods in the ``put the teacup into the box'' task. It first segments the task into multiple atomic subtasks and designs corresponding atomic state/action spaces for exploring a more robust data collection process.
 }\label{fig:method}
\end{figure*}

\textbf{HD-Space: Atomic tasks Segmentation.} Given a long-horizon manipulation task, it initiates by pinpointing the nodes of key actions, taking into account the crucial manipulation details.
As shown in Fig.~\ref{fig:overall} (c) and Fig.~\ref{fig:method}, it segments \texttt{put the teacup into the box} task into four semantically meaningful subtasks: \texttt{open box}, \texttt{pick up cup}, \texttt{put in box}, and \texttt{close box}. The basis of atomic action segmentation is the moment from the need to locate a certain pose to the execution of fine operations.  For example, the first atomic action is that the end of the robotic arm moves to the rope of the box, and then the precise operation content is to close the gripper and open the box lid.

\textbf{HD-Space: Atomic Data Collection Space.} After segmenting all atomic tasks, we elaborate on the design of the data collection space for each atomic task. Intuitively, in order for the policy to drive the robotic arm to the target point and perform fine operations, it is necessary to allow the model to see as many states around the target point as possible. To this end, we \emph{evenly} placed the end of the robotic arm around the target point and then took this state as the starting point for human demonstration and data recording until the completion of an atomic task. Taking the \texttt{put the teacup into the box} task in Fig.~\ref{fig:method} of the green region as an example: Atomic Space$\#1$: Place the end of the robotic arm in the space around the box, then move it to the box rope, close the gripper and open the box; Atomic Space$\#2$: Place the end of the robotic arm in the space around the cup, then move it to the edge of the cup mouth, close the gripper and move to the top of the box; Atomic Space$\#3$: The robotic arm holds the cup and hovers evenly around the top of the box as the starting point, then moves to the inside of the box, opens the gripper and lowers the cup. Atomic Space$\#4$: Place the end of the robotic arm in the space around the box, then move it to the box rope, close the gripper, and close the box. 

\begin{figure*}[t]
    \centering
    \includegraphics[width=1.0\linewidth]{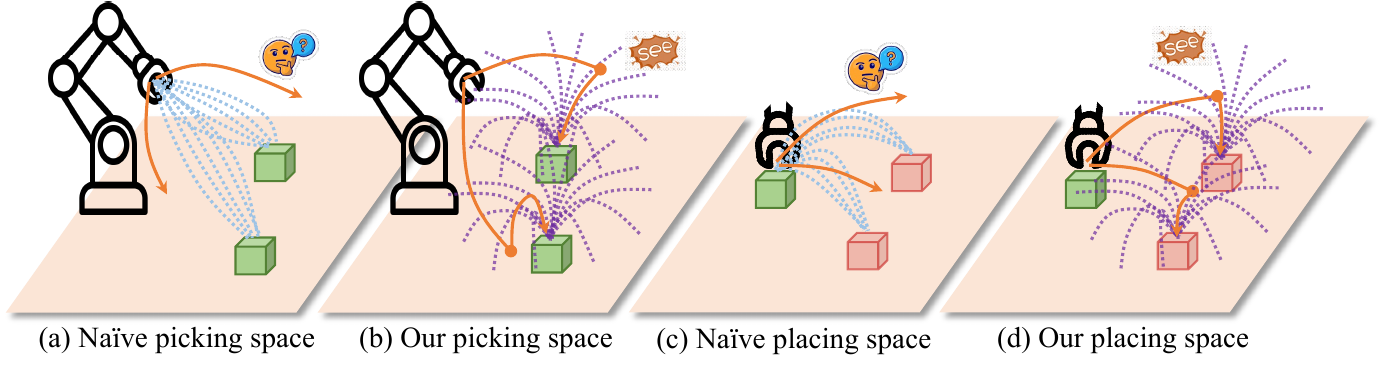}
    \vspace{-0.6cm}
    \caption{\textbf{Robust space considerations}. Blue and purple dotted lines represent the trajectories covered by naive and HD-Space data collection processes. Orange lines represent prediction actions.
 }\label{fig:space}
\end{figure*}

\textbf{Practical Considerations.} As shown in Fig~\ref{fig:space}, we further analyze the robustness of HD-Space.
It has seen more states around the fine manipulation location by uniformly traversing the state/action space of potential trajectory error predictions for each atomic task, HD-Space can proactively suppress the erroneous prediction trajectories from the error-prone space.
On the contrary, the naive data collection space only starts from a fixed end-effector pose and moves toward the target object. In this case, the collected data only passes through a narrow state space. When the trajectory predicted by the model deviates from this distribution, it is very easy to make wrong decisions, thereby causing the failure of the manipulation task.

\textbf{Data Usage.} We use the representative action chunking with transformers (ACT) model~\citep{zhao2023learning} to evaluate the performance of HD-Space.
It uses the transformer-based encoder and decoder architecture the same as \citep{carion2020end}. The decoder interacts with the image features using randomly initialized queries and then outputs a chunk of actions of size. On this basis, we discard the conditional VAE~\citep{sohn2015learning, kingma2013auto} of ACT. To enhance the perception capabilities, we use the DINOv2-small~\citep{oquab2023dinov2} as the image feature extractor instead of ResNet-18, which has been proven effective in the IL model~\citep{li2024virt,borucka2019city}. In addition, we used an MAE-based data augmentation inspired from ~\citep{he2022masked} for the input images during training, which divides the image into uniform small square pieces and then randomly replaces the squares with zero values. During the training, we focused on exploring the impact of the ratio between the data collected by the original method and the HD-Space method on performance and efficiency.

\section{Experimental Results}
\label{sec:result}

We formulate our experiments to answer the following questions: \textbf{Q1:} Can HD-Space steadily improve the performance of the policy? \textbf{Q2:} Can HD-Space enable the model to complete tasks with a longer horizon? \textbf{Q3:} Can HD-Space reduce the cost of human demonstration for IL?

\subsection{Implementation Details}

\textbf{Robots.} We leverage two representative robots for the human demonstration process and policy deployment: 1) the Cobot Mobile ALOHA using the Mobile ALOHA system design~\citep{fu2024mobile}; 2) the piper arm robots with isomorphic teleoperation system designed by agilex.ai. In addition, we use the ViperX 6dof robot~\citep{Viperx} to assess our stronger policy on simulated fine manipulation tasks, which follow the same setting with ALOHA~\citep{zhao2023learning}.

\begin{figure*}[h]
    \centering
    \includegraphics[width=0.94\linewidth]{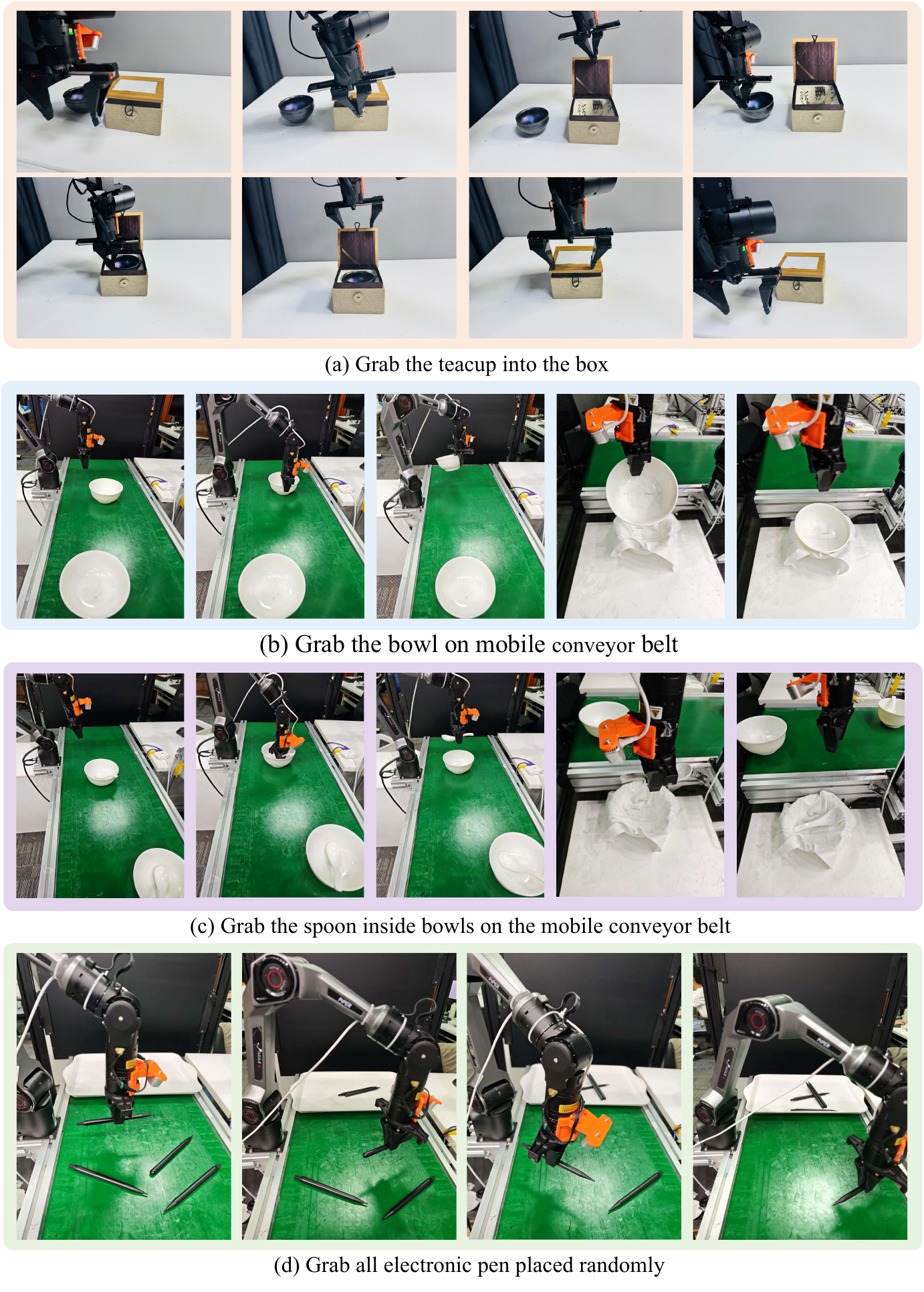}
    \vspace{-0.4cm}
    \caption{\textbf{Long-horizon tasks} including one desktop task and two types of factory-related tasks.
 }\label{fig:task}
\end{figure*}

\textbf{Tasks.}
We assess the performance of HD-Space in three real-world long-horizon tasks with human demonstrations, including one desktop task and two factory-related tasks:
\textbf{1) Grabbing the Teacup into the Box:} The robotic arm needs to open the box first, then pick up the tea cup and put it into the box, and finally close the box (see Fig.~\ref{fig:task} (a));
\textbf{2) Grabbing Mobile objects:} The robotic arm needs to grab objects that appear on the conveyor belt at any time, \emph{e.g.,} bowls and spoons in bowls. Grabbing the bowls only needs dynamic position generalization, while grabbing spoons requires an additional generalization of the placement angle of the spoon and does not take the bowls away together. The conveyor belt can be set with different moving speed challenges~\citep{streaming};
\textbf{3) Grabbing unordered objects:} The robotic arm needs to pick up all the electronic pens on the table that are randomly squared and then place them in the storage.
To our knowledge, this is the first work to challenge \textbf{mobile and unordered object manipulation with an end-to-end imitation learning model}.

\begin{table}[h]
    \small
    \begin{center}\renewcommand{\arraystretch}{1.01}
        \vspace{-0.3cm}
        \resizebox{1.0\linewidth}{!}{
		\begin{tabular}{lcc}
        \toprule
            \textbf{Method} & \textbf{Transfer Cube (Sim)} & \textbf{Peg Insertion (Sim)} \\
            \midrule
            ACT~\citep{zhao2023learning} & 50 & 20\\
            ACT+DINOv2~\citep{oquab2023dinov2} & 67 & 33 \\
            \textbf{Ours: ACT+DINOv2~\citep{oquab2023dinov2}+MAE-Augmentation} & \textbf{73} & \textbf{36}\\
            \bottomrule
        \end{tabular}
     }
    \end{center}
    \caption{\textbf{Effectiveness of stronger IL model}. Following ACT~\citep{zhao2023learning}, we use success rate ($\%$) for task evaluation and average the results across 3 random seeds over 50 episodes each.}\label{table:stronger_model}
    \vspace{-0.2in}
\end{table}

\textbf{Effectiveness of the Stronger Model.} The improvement results are summarized in Tab.~\ref{table:stronger_model}. Our IL model significantly improves performance compared to the baseline ACT model on the \texttt{Transfer Cube} and \texttt{Peg Insertion} simulated bimanual Manipulation tasks. It shows that the DINOv2 and MAE augmentation can enhance the precision of fine manipulation tasks.
This high-performance advantage provides a better model infrastructure to evaluate the performance of the HD-Space data collection method.

\subsection{Experiment Results}

In this subsection, we conduct three quantitative experiments to answer the performance improvement, completing more longer sequence tasks, and reducing the data collection cost, respectively.

\begin{figure*}[h]
    \centering
    \includegraphics[width=1.0\linewidth]{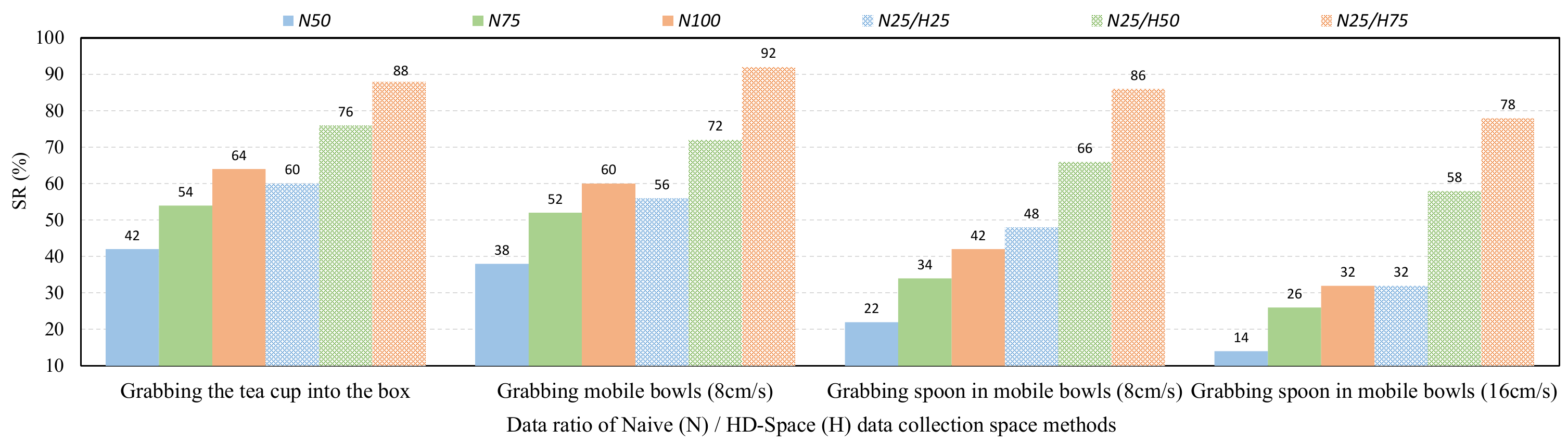}
    \vspace{-0.6cm}
    \caption{\textbf{Ablation studies of success rate (SR)} on 3 real-scene tasks with different data episodes. 
 }\label{fig:ablation_sr}
\end{figure*}

\textbf{A1: Performance Improvement.}
As shown in Fig.~\ref{fig:ablation_sr}, when the episode numbers of data are strictly equal, the HD-Space method demonstrates significant performance improvements. It consistently improves over naive data collection method: boosting 8\%, 8\%, and 8\% SR on the \texttt{Grabbing the teacup into the box } task; improving 18\%, 20\%, and 32\% SR on the \texttt{Grabbing mobile bowls} task; enhancing 20\%, 32\%, and 44\% SR on the \texttt{Grabbing spoon in mobile bowls} task, respectively. Especially in grabbing mobile objects tasks with different mobile speeds, the performance of HD-Space using data with fewer episode numbers has exceeded that of the naive method with more episode numbers. For example, using 25 naive and 25 HD-Space episodes can outperform the 100 episodes of the naive method. It reveals that HD-Space is suitable for enhancing tasks with higher requirements for position, angle, and speed. In addition, by adding the data of HD-Space, the increase in performance is much faster than that of the original method, which may provide new insight for the data calling efforts in the robotic learning field and technology application.

\begin{table}[h]
    \centering
    \resizebox{1.0\linewidth}{!}{
    \begin{tabular}{lcccccc}
    \toprule
     & \multicolumn{2}{c}{\textbf{50 episodes}} & \multicolumn{2}{c}{\textbf{75 episodes}} & \multicolumn{2}{c}{\textbf{100 episodes}}\\
    \cmidrule(lr){2-3}
    \cmidrule(lr){4-5}
    \cmidrule(lr){6-7}
     & N50 & N25+H25 & N75 & N25+H50 & N100 & N25+H75 \\
    \midrule
    \textbf{Grab all electronic placed randomly (Max lenght=5)} & 2.34 & \textbf{3.18} (+0.84) & 2.76 & \textbf{3.91} (+1.15) & 3.23 & \textbf{4.45} (+1.22) \\
    \textbf{Grab the tea cup into the box (Max lenght=4)}  & 2.84 & \textbf{3.16} (+0.32) & 3.08 & \textbf{3.48} (+0.40) & 3.24 & \textbf{3.78} (+0.54) \\
    \bottomrule
    \end{tabular}}
    \vspace{0.2cm}
    \caption{
    \textbf{Ablation studies of the average number of completed sequential tasks} on the grabbing unordered objects and the putting teacup real-scene tasks with different data episode ratios.
    }\label{tab:assl}
\vspace{-0.4cm}
\end{table}

\textbf{A2: Complete Longer Sequence Tasks.}
We conduct these quantitative experiments on the task randomly and without stacking five electronic pens on the plane, and then the robot arm picks up electronic pens one by one and places them in the fixed plates, which report the grasp number of pens. In addition, we also report the complete substak numbers of the \texttt{put teacup into box} task.
As shown in Tab.~\ref{tab:assl}, we report the average number of completed sequential task evaluation indicators. The same as Fig.~\ref{fig:ablation_sr}, in the case of the same episodes of training data, HD-Space can achieve more average number of completed sequential tasks. With the increase of HD-Space data, the increase in performance has also maintained a strong growth, which shows that our proposed method enables IL model to cover a longer-horizon task better.

\begin{table}[h]
    \centering
    \resizebox{0.96\linewidth}{!}{
    \begin{tabular}{lcccccc}
    \toprule
     & \textbf{Grab the Teacup} & \textbf{Grab mobile bowl} & \textbf{Grab mobile spoon} & \textbf{Grab electronic pens}  \\
    \midrule
    \textbf{Naive}    & 600  & 350 & 450 & 1300 \\
    \textbf{HD-Space} & \textbf{400}  & \textbf{100} & \textbf{120} & \textbf{120}  \\
    \bottomrule
    \end{tabular}}
    \vspace{0.2cm}
    \caption{
    \textbf{Data collection cost} comparison between naive and HD-Space schemes. We record the total number of frames in each episode.
    }\label{tab:data_collection_cost}
\vspace{-0.4cm}
\end{table}

\textbf{A3: Reduce the Data Collection Cost.}
As shown in Tab.~\ref{tab:data_collection_cost}, we compare the number of frames of each episode between naive and our HD-Space data collection methods. It demonstrates that HD-Space uses fewer image frames to generate data since our method only needs to record the segments that are prone to compound trajectory errors or important actions. Like in \texttt{Grab mobile spoon} task, it only needs to record a short section of the process where the robot end effector moves from around the spoon to above it, and then rotates at an appropriate Angle downward to grasp. Combined with the results in Fig.~\ref{fig:ablation_sr}, HD-Space enables utilize less data episodes and frames per episode to train a better model. In addition, segmenting the task into hierarchical atomic spaces for data collection is more conducive to experts collecting consistent data with less noise, thereby further reducing practical costs and improving data quality.

\begin{figure*}[h]
    \centering
    \includegraphics[width=1.0\linewidth]{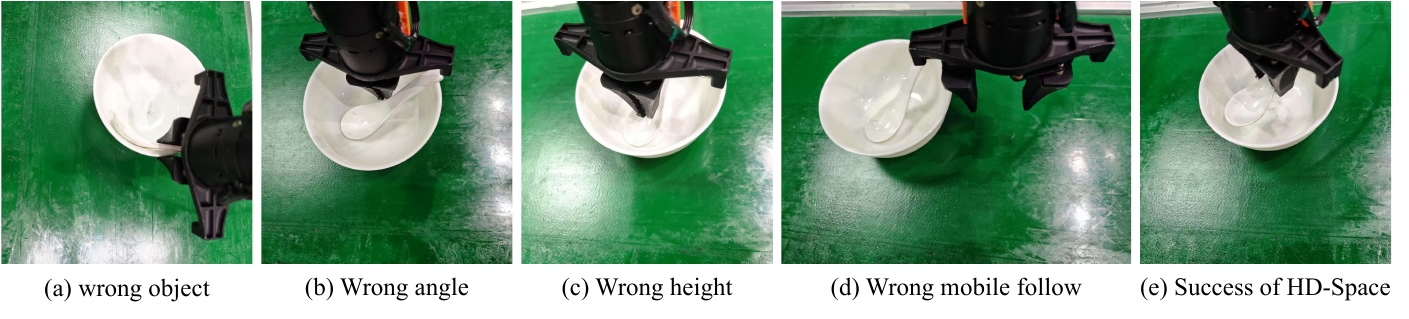}
    \vspace{-0.7cm}
    \caption{\textbf{Qualitative analysis} of the failed cases and the successful ones solved by HD-Space.
 }\label{fig:qualitative_analysis}
\vspace{-0.2cm}
\end{figure*}

\subsection{Qualitative Analysis.}

We monitored multiple erroneous operation behaviors on the \texttt{Grabbing spoon in mobile bowls} task. As shown in Fig.~\ref{fig:qualitative_analysis}, we categorize the failure cases into 4 types: 1) mistakenly picking up the bowl along with grabbing the spoon; 2) predicting the wrong Angle of the end effector, resulting in the inability to accurately grasp the spoon; 3) predicting incorrectly height, resulting in spoon grabbed empty-handed; 4) failing to follow the mobile object, leading to the inability to grab a spoon. Adding more HD-Space data for IL training can solve the above problems, which shows that our method can enhance the active perception ability and avoid the compound trajectory errors of unseen states.

\section{Conclusion}
\label{sec:conclusion}

We present HD-Space, a hierarchical data collection space to collect high-quality data for training a better model with imitation learning. Our method achieves significant performance improvement in three challenging manipulation tasks. Especially, it can make the IL model conduct more sequential subtasks with fewer collected data and human demonstration costs. HD-Space provides new insight into robotic imitation learning, which is expected to guide a new effort for future IL data scaling.

\textbf{Limitation and Future Work.} 
HD-Space only designs a better data collection scheme for a single-task model to learn a fine manipulation task. In future work, HD-Space can be implanted into multi-task and vision-language-action (VLA) models, which may be interesting to design HD-Space with sharing the same condition of text description or image goal.

\clearpage

\bibliography{example}  

\begin{thebibliography}{62}
\providecommand{\natexlab}[1]{#1}
\providecommand{\url}[1]{\texttt{#1}}
\expandafter\ifx\csname urlstyle\endcsname\relax
  \providecommand{\doi}[1]{doi: #1}\else
  \providecommand{\doi}{doi: \begingroup \urlstyle{rm}\Url}\fi

\bibitem[Pomerleau(1988)]{pomerleau1988alvinn}
D.~A. Pomerleau.
\newblock Alvinn: An autonomous land vehicle in a neural network.
\newblock \emph{Advances in neural information processing systems}, 1, 1988.

\bibitem[Hussein et~al.(2017)Hussein, Gaber, Elyan, and Jayne]{hussein2017imitation}
A.~Hussein, M.~M. Gaber, E.~Elyan, and C.~Jayne.
\newblock Imitation learning: A survey of learning methods.
\newblock \emph{ACM Computing Surveys (CSUR)}, 50\penalty0 (2):\penalty0 1--35, 2017.

\bibitem[Jang et~al.(2022)Jang, Irpan, Khansari, Kappler, Ebert, Lynch, Levine, and Finn]{jang2022bc}
E.~Jang, A.~Irpan, M.~Khansari, D.~Kappler, F.~Ebert, C.~Lynch, S.~Levine, and C.~Finn.
\newblock Bc-z: Zero-shot task generalization with robotic imitation learning.
\newblock In \emph{Conference on Robot Learning}, pages 991--1002. PMLR, 2022.

\bibitem[Mandlekar et~al.(2021)Mandlekar, Xu, Wong, Nasiriany, Wang, Kulkarni, Fei-Fei, Savarese, Zhu, and Mart{\'\i}n-Mart{\'\i}n]{mandlekar2021matters}
A.~Mandlekar, D.~Xu, J.~Wong, S.~Nasiriany, C.~Wang, R.~Kulkarni, L.~Fei-Fei, S.~Savarese, Y.~Zhu, and R.~Mart{\'\i}n-Mart{\'\i}n.
\newblock What matters in learning from offline human demonstrations for robot manipulation.
\newblock \emph{Conference on Robot Learning (CoRL)}, 2021.

\bibitem[Zhao et~al.(2023)Zhao, Kumar, Levine, and Finn]{zhao2023learning}
T.~Z. Zhao, V.~Kumar, S.~Levine, and C.~Finn.
\newblock Learning fine-grained bimanual manipulation with low-cost hardware.
\newblock \emph{arXiv preprint arXiv:2304.13705}, 2023.

\bibitem[Chi et~al.(2023)Chi, Xu, Feng, Cousineau, Du, Burchfiel, Tedrake, and Song]{chi2023diffusion}
C.~Chi, Z.~Xu, S.~Feng, E.~Cousineau, Y.~Du, B.~Burchfiel, R.~Tedrake, and S.~Song.
\newblock Diffusion policy: Visuomotor policy learning via action diffusion.
\newblock \emph{The International Journal of Robotics Research}, page 02783649241273668, 2023.

\bibitem[Ze et~al.(2024)Ze, Zhang, Zhang, Hu, Wang, and Xu]{ze20243d}
Y.~Ze, G.~Zhang, K.~Zhang, C.~Hu, M.~Wang, and H.~Xu.
\newblock 3d diffusion policy: Generalizable visuomotor policy learning via simple 3d representations.
\newblock \emph{arXiv preprint arXiv:2403.03954}, 2024.

\bibitem[Goyal et~al.(2023)Goyal, Xu, Guo, Blukis, Chao, and Fox]{goyal2023rvt}
A.~Goyal, J.~Xu, Y.~Guo, V.~Blukis, Y.-W. Chao, and D.~Fox.
\newblock Rvt: Robotic view transformer for 3d object manipulation.
\newblock In \emph{Conference on Robot Learning}, pages 694--710. PMLR, 2023.

\bibitem[Chi et~al.(2024)Chi, Xu, Pan, Cousineau, Burchfiel, Feng, Tedrake, and Song]{umi}
C.~Chi, Z.~Xu, C.~Pan, E.~Cousineau, B.~Burchfiel, S.~Feng, R.~Tedrake, and S.~Song.
\newblock Universal manipulation interface: In-the-wild robot teaching without in-the-wild robots.
\newblock \emph{arXiv preprint arXiv:2402.10329}, 2024.

\bibitem[Fu et~al.(2024{\natexlab{a}})Fu, Zhao, and Finn]{fu2024mobile}
Z.~Fu, T.~Z. Zhao, and C.~Finn.
\newblock Mobile aloha: Learning bimanual mobile manipulation with low-cost whole-body teleoperation.
\newblock \emph{arXiv preprint arXiv:2401.02117}, 2024{\natexlab{a}}.

\bibitem[Fu et~al.(2024{\natexlab{b}})Fu, Zhao, Wu, Wetzstein, and Finn]{fu2024humanplus}
Z.~Fu, Q.~Zhao, Q.~Wu, G.~Wetzstein, and C.~Finn.
\newblock Humanplus: Humanoid shadowing and imitation from humans.
\newblock \emph{arXiv preprint arXiv:2406.10454}, 2024{\natexlab{b}}.

\bibitem[Cheng et~al.(2024)Cheng, Li, Yang, Yang, and Wang]{cheng2024open}
X.~Cheng, J.~Li, S.~Yang, G.~Yang, and X.~Wang.
\newblock Open-television: Teleoperation with immersive active visual feedback.
\newblock \emph{arXiv preprint arXiv:2407.01512}, 2024.

\bibitem[Lin et~al.(2024)Lin, Hu, Sheng, Wen, You, and Gao]{lin2024data}
F.~Lin, Y.~Hu, P.~Sheng, C.~Wen, J.~You, and Y.~Gao.
\newblock Data scaling laws in imitation learning for robotic manipulation.
\newblock \emph{arXiv preprint arXiv:2410.18647}, 2024.

\bibitem[Lillicrap et~al.(2015)Lillicrap, Hunt, Pritzel, Heess, Erez, Tassa, Silver, and Wierstra]{lillicrap2015continuous}
T.~P. Lillicrap, J.~J. Hunt, A.~Pritzel, N.~Heess, T.~Erez, Y.~Tassa, D.~Silver, and D.~Wierstra.
\newblock Continuous control with deep reinforcement learning.
\newblock \emph{arXiv preprint arXiv:1509.02971}, 2015.

\bibitem[Tan et~al.(2024)Tan, Xu, Ying, Mao, Liu, Zhang, Su, and Zhu]{tan2024manibox}
H.~Tan, X.~Xu, C.~Ying, X.~Mao, S.~Liu, X.~Zhang, H.~Su, and J.~Zhu.
\newblock Manibox: Enhancing spatial grasping generalization via scalable simulation data generation.
\newblock \emph{arXiv preprint arXiv:2411.01850}, 2024.

\bibitem[Mandlekar et~al.(2020)Mandlekar, Xu, Mart{\'\i}n-Mart{\'\i}n, Zhu, Fei-Fei, and Savarese]{mandlekar2020human}
A.~Mandlekar, D.~Xu, R.~Mart{\'\i}n-Mart{\'\i}n, Y.~Zhu, L.~Fei-Fei, and S.~Savarese.
\newblock Human-in-the-loop imitation learning using remote teleoperation.
\newblock \emph{arXiv preprint arXiv:2012.06733}, 2020.

\bibitem[Luo et~al.(2024)Luo, Xu, Wu, and Levine]{luo2024precise}
J.~Luo, C.~Xu, J.~Wu, and S.~Levine.
\newblock Precise and dexterous robotic manipulation via human-in-the-loop reinforcement learning.
\newblock \emph{arXiv preprint arXiv:2410.21845}, 2024.

\bibitem[Hiranaka et~al.(2023)Hiranaka, Hwang, Lee, Wang, Fei-Fei, Wu, and Zhang]{hiranaka2023primitive}
A.~Hiranaka, M.~Hwang, S.~Lee, C.~Wang, L.~Fei-Fei, J.~Wu, and R.~Zhang.
\newblock Primitive skill-based robot learning from human evaluative feedback.
\newblock In \emph{2023 IEEE/RSJ International Conference on Intelligent Robots and Systems (IROS)}, pages 7817--7824. IEEE, 2023.

\bibitem[Li et~al.(2024)Li, Li, Zhang, Zhang, Jia, Wang, Fan, Tseng, and Wang]{li2024robogsim}
X.~Li, J.~Li, Z.~Zhang, R.~Zhang, F.~Jia, T.~Wang, H.~Fan, K.-K. Tseng, and R.~Wang.
\newblock Robogsim: A real2sim2real robotic gaussian splatting simulator.
\newblock \emph{arXiv preprint arXiv:2411.11839}, 2024.

\bibitem[Katara et~al.(2024)Katara, Xian, and Fragkiadaki]{katara2024gen2sim}
P.~Katara, Z.~Xian, and K.~Fragkiadaki.
\newblock Gen2sim: Scaling up robot learning in simulation with generative models.
\newblock In \emph{2024 IEEE International Conference on Robotics and Automation (ICRA)}, pages 6672--6679. IEEE, 2024.

\bibitem[Wang et~al.(2023)Wang, Johnson, Lu, Huang, Booth, Kramer-Bottiglio, Aanjaneya, and Bekris]{wang2023real2sim2real1}
K.~Wang, W.~R. Johnson, S.~Lu, X.~Huang, J.~Booth, R.~Kramer-Bottiglio, M.~Aanjaneya, and K.~Bekris.
\newblock Real2sim2real transfer for control of cable-driven robots via a differentiable physics engine.
\newblock In \emph{2023 IEEE/RSJ International Conference on Intelligent Robots and Systems (IROS)}, pages 2534--2541. IEEE, 2023.

\bibitem[Wu et~al.(2024)Wu, Pan, Wu, Wang, Miao, Xu, and Wang]{wu2024rl}
Y.~Wu, L.~Pan, W.~Wu, G.~Wang, Y.~Miao, F.~Xu, and H.~Wang.
\newblock Rl-gsbridge: 3d gaussian splatting based real2sim2real method for robotic manipulation learning.
\newblock \emph{arXiv preprint arXiv:2409.20291}, 2024.

\bibitem[Han et~al.(2025)Han, Liu, Chen, Yu, Lyu, Tian, Wang, Zhang, and Pang]{han2025re}
X.~Han, M.~Liu, Y.~Chen, J.~Yu, X.~Lyu, Y.~Tian, B.~Wang, W.~Zhang, and J.~Pang.
\newblock Re3sim: Generating high-fidelity simulation data via 3d-photorealistic real-to-sim for robotic manipulation.
\newblock \emph{arXiv preprint arXiv:2502.08645}, 2025.

\bibitem[Wang et~al.(2023)Wang, Guo, Vuong, Qin, Su, and Christensen]{wang2023real2sim2real}
L.~Wang, R.~Guo, Q.~Vuong, Y.~Qin, H.~Su, and H.~Christensen.
\newblock A real2sim2real method for robust object grasping with neural surface reconstruction.
\newblock In \emph{2023 IEEE 19th International Conference on Automation Science and Engineering (CASE)}, pages 1--8. IEEE, 2023.

\bibitem[Zhang et~al.(2023)Zhang, Wang, Sun, Wu, Zhu, and Tomizuka]{zhang2023efficient}
X.~Zhang, C.~Wang, L.~Sun, Z.~Wu, X.~Zhu, and M.~Tomizuka.
\newblock Efficient sim-to-real transfer of contact-rich manipulation skills with online admittance residual learning.
\newblock In \emph{Conference on Robot Learning (CoRL)}, pages 1621--1639. PMLR, 2023.

\bibitem[Schaal(1999)]{schaal1999imitation}
S.~Schaal.
\newblock Is imitation learning the route to humanoid robots?
\newblock \emph{Trends in cognitive sciences}, 3\penalty0 (6):\penalty0 233--242, 1999.

\bibitem[Argall et~al.(2009)Argall, Chernova, Veloso, and Browning]{argall2009survey}
B.~D. Argall, S.~Chernova, M.~Veloso, and B.~Browning.
\newblock A survey of robot learning from demonstration.
\newblock \emph{Robotics and autonomous systems}, 57\penalty0 (5):\penalty0 469--483, 2009.

\bibitem[Shi et~al.(2023)Shi, Sharma, Zhao, and Finn]{shi2023waypoint}
L.~X. Shi, A.~Sharma, T.~Z. Zhao, and C.~Finn.
\newblock Waypoint-based imitation learning for robotic manipulation.
\newblock \emph{Conference on Robot Learning (CoRL)}, 2023.

\bibitem[James and Davison(2022)]{james2022q}
S.~James and A.~J. Davison.
\newblock Q-attention: Enabling efficient learning for vision-based robotic manipulation.
\newblock \emph{IEEE Robotics and Automation Letters}, 7\penalty0 (2):\penalty0 1612--1619, 2022.

\bibitem[Pari et~al.(2021)Pari, Shafiullah, Arunachalam, and Pinto]{pari2021surprising}
J.~Pari, N.~M. Shafiullah, S.~P. Arunachalam, and L.~Pinto.
\newblock The surprising effectiveness of representation learning for visual imitation.
\newblock \emph{arXiv preprint arXiv:2112.01511}, 2021.

\bibitem[Shafiullah et~al.(2022)Shafiullah, Cui, Altanzaya, and Pinto]{shafiullah2022behavior}
N.~M. Shafiullah, Z.~Cui, A.~A. Altanzaya, and L.~Pinto.
\newblock Behavior transformers: Cloning $ k $ modes with one stone.
\newblock \emph{Advances in neural information processing systems}, 35:\penalty0 22955--22968, 2022.

\bibitem[Zhu et~al.(2023)Zhu, Joshi, Stone, and Zhu]{zhu2023viola}
Y.~Zhu, A.~Joshi, P.~Stone, and Y.~Zhu.
\newblock Viola: Imitation learning for vision-based manipulation with object proposal priors.
\newblock In \emph{Conference on Robot Learning}, pages 1199--1210. PMLR, 2023.

\bibitem[Lee et~al.(2024)Lee, Chuang, Chen, and Soltani]{lee2024interact}
A.~Lee, I.~Chuang, L.-Y. Chen, and I.~Soltani.
\newblock Interact: Inter-dependency aware action chunking with hierarchical attention transformers for bimanual manipulation.
\newblock \emph{Conference on Robot Learning (CoRL)}, 2024.

\bibitem[Chen et~al.(2025)Chen, Yuan, Mu, and Su]{chen2025responsive}
Z.~Chen, X.~Yuan, T.~Mu, and H.~Su.
\newblock Responsive noise-relaying diffusion policy: Responsive and efficient visuomotor control.
\newblock \emph{arXiv preprint arXiv:2502.12724}, 2025.

\bibitem[Pearce et~al.(2023)Pearce, Rashid, Kanervisto, Bignell, Sun, Georgescu, Macua, Tan, Momennejad, Hofmann, et~al.]{pearce2023imitating}
T.~Pearce, T.~Rashid, A.~Kanervisto, D.~Bignell, M.~Sun, R.~Georgescu, S.~V. Macua, S.~Z. Tan, I.~Momennejad, K.~Hofmann, et~al.
\newblock Imitating human behaviour with diffusion models.
\newblock \emph{arXiv preprint arXiv:2301.10677}, 2023.

\bibitem[Bharadhwaj et~al.(2024)Bharadhwaj, Vakil, Sharma, Gupta, Tulsiani, and Kumar]{bharadhwaj2024roboagent}
H.~Bharadhwaj, J.~Vakil, M.~Sharma, A.~Gupta, S.~Tulsiani, and V.~Kumar.
\newblock Roboagent: Generalization and efficiency in robot manipulation via semantic augmentations and action chunking.
\newblock In \emph{2024 IEEE International Conference on Robotics and Automation (ICRA)}, pages 4788--4795. IEEE, 2024.

\bibitem[Haldar et~al.(2024)Haldar, Peng, and Pinto]{haldar2024baku}
S.~Haldar, Z.~Peng, and L.~Pinto.
\newblock Baku: An efficient transformer for multi-task policy learning.
\newblock \emph{arXiv preprint arXiv:2406.07539}, 2024.

\bibitem[Brohan et~al.(2022)Brohan, Brown, Carbajal, Chebotar, Dabis, Finn, Gopalakrishnan, Hausman, Herzog, Hsu, et~al.]{brohan2022rt}
A.~Brohan, N.~Brown, J.~Carbajal, Y.~Chebotar, J.~Dabis, C.~Finn, K.~Gopalakrishnan, K.~Hausman, A.~Herzog, J.~Hsu, et~al.
\newblock Rt-1: Robotics transformer for real-world control at scale.
\newblock \emph{arXiv preprint arXiv:2212.06817}, 2022.

\bibitem[Kim et~al.(2024)Kim, Pertsch, Karamcheti, Xiao, Balakrishna, Nair, Rafailov, Foster, Lam, Sanketi, et~al.]{kim2024openvla}
M.~J. Kim, K.~Pertsch, S.~Karamcheti, T.~Xiao, A.~Balakrishna, S.~Nair, R.~Rafailov, E.~Foster, G.~Lam, P.~Sanketi, et~al.
\newblock Openvla: An open-source vision-language-action model.
\newblock \emph{arXiv preprint arXiv:2406.09246}, 2024.

\bibitem[Black et~al.(2024)Black, Brown, Driess, Esmail, Equi, Finn, Fusai, Groom, Hausman, Ichter, et~al.]{black2024pi_0}
K.~Black, N.~Brown, D.~Driess, A.~Esmail, M.~Equi, C.~Finn, N.~Fusai, L.~Groom, K.~Hausman, B.~Ichter, et~al.
\newblock $pi0$: A vision-language-action flow model for general robot control.
\newblock \emph{arXiv preprint arXiv:2410.24164}, 2024.

\bibitem[O’Neill et~al.(2024)O’Neill, Rehman, Maddukuri, Gupta, Padalkar, Lee, Pooley, Gupta, Mandlekar, Jain, et~al.]{o2024open}
A.~O’Neill, A.~Rehman, A.~Maddukuri, A.~Gupta, A.~Padalkar, A.~Lee, A.~Pooley, A.~Gupta, A.~Mandlekar, A.~Jain, et~al.
\newblock Open x-embodiment: Robotic learning datasets and rt-x models: Open x-embodiment collaboration 0.
\newblock In \emph{2024 IEEE International Conference on Robotics and Automation (ICRA)}, pages 6892--6903. IEEE, 2024.

\bibitem[Khazatsky et~al.(2024)Khazatsky, Pertsch, Nair, Balakrishna, Dasari, Karamcheti, Nasiriany, Srirama, Chen, Ellis, et~al.]{khazatsky2024droid}
A.~Khazatsky, K.~Pertsch, S.~Nair, A.~Balakrishna, S.~Dasari, S.~Karamcheti, S.~Nasiriany, M.~K. Srirama, L.~Y. Chen, K.~Ellis, et~al.
\newblock Droid: A large-scale in-the-wild robot manipulation dataset.
\newblock \emph{arXiv preprint arXiv:2403.12945}, 2024.

\bibitem[Wu et~al.(2024)Wu, Hou, Liu, Che, Ju, Yang, Li, Zhao, Xu, Yang, et~al.]{wu2024robomind}
K.~Wu, C.~Hou, J.~Liu, Z.~Che, X.~Ju, Z.~Yang, M.~Li, Y.~Zhao, Z.~Xu, G.~Yang, et~al.
\newblock Robomind: Benchmark on multi-embodiment intelligence normative data for robot manipulation.
\newblock \emph{arXiv preprint arXiv:2412.13877}, 2024.

\bibitem[Shridhar et~al.(2023)Shridhar, Manuelli, and Fox]{shridhar2023perceiver}
M.~Shridhar, L.~Manuelli, and D.~Fox.
\newblock Perceiver-actor: A multi-task transformer for robotic manipulation.
\newblock In \emph{Conference on Robot Learning}, pages 785--799. PMLR, 2023.

\bibitem[Hsiao and Lozano-Perez(2006)]{hsiao2006imitation}
K.~Hsiao and T.~Lozano-Perez.
\newblock Imitation learning of whole-body grasps.
\newblock In \emph{2006 IEEE/RSJ international conference on intelligent robots and systems}, pages 5657--5662. IEEE, 2006.

\bibitem[Akgun et~al.(2012)Akgun, Cakmak, Jiang, and Thomaz]{akgun2012keyframe}
B.~Akgun, M.~Cakmak, K.~Jiang, and A.~L. Thomaz.
\newblock Keyframe-based learning from demonstration: Method and evaluation.
\newblock \emph{International Journal of Social Robotics}, 4:\penalty0 343--355, 2012.

\bibitem[Zeng et~al.(2021)Zeng, Florence, Tompson, Welker, Chien, Attarian, Armstrong, Krasin, Duong, Sindhwani, et~al.]{zeng2021transporter}
A.~Zeng, P.~Florence, J.~Tompson, S.~Welker, J.~Chien, M.~Attarian, T.~Armstrong, I.~Krasin, D.~Duong, V.~Sindhwani, et~al.
\newblock Transporter networks: Rearranging the visual world for robotic manipulation.
\newblock In \emph{Conference on Robot Learning}, pages 726--747. PMLR, 2021.

\bibitem[Shridhar et~al.(2022)Shridhar, Manuelli, and Fox]{shridhar2022cliport}
M.~Shridhar, L.~Manuelli, and D.~Fox.
\newblock Cliport: What and where pathways for robotic manipulation.
\newblock In \emph{Conference on robot learning}, pages 894--906. PMLR, 2022.

\bibitem[Morrison et~al.(2018)Morrison, Corke, and Leitner]{morrison2018closing}
D.~Morrison, P.~Corke, and J.~Leitner.
\newblock Closing the loop for robotic grasping: A real-time, generative grasp synthesis approach.
\newblock \emph{arXiv preprint arXiv:1804.05172}, 2018.

\bibitem[Huang et~al.(2025)Huang, Liao, Feng, Jiang, Liu, Li, Yao, and Ren]{huang2025adversarial}
S.~Huang, Y.~Liao, S.~Feng, S.~Jiang, S.~Liu, H.~Li, M.~Yao, and G.~Ren.
\newblock Adversarial data collection: Human-collaborative perturbations for efficient and robust robotic imitation learning.
\newblock \emph{arXiv preprint arXiv:2503.11646}, 2025.

\bibitem[Belkhale et~al.(2023)Belkhale, Cui, and Sadigh]{belkhale2023data}
S.~Belkhale, Y.~Cui, and D.~Sadigh.
\newblock Data quality in imitation learning.
\newblock \emph{Advances in neural information processing systems}, 36:\penalty0 80375--80395, 2023.

\bibitem[Samadi et~al.(2024)Samadi, Koufos, Debattista, and Dianati]{samadi2024good}
A.~Samadi, K.~Koufos, K.~Debattista, and M.~Dianati.
\newblock Good data is all imitation learning needs.
\newblock \emph{arXiv preprint arXiv:2409.17605}, 2024.

\bibitem[Lin et~al.(2024)Lin, Hu, Sheng, Wen, You, and Gao]{lin2024data1}
F.~Lin, Y.~Hu, P.~Sheng, C.~Wen, J.~You, and Y.~Gao.
\newblock Data scaling laws in imitation learning for robotic manipulation.
\newblock \emph{arXiv preprint arXiv:2410.18647}, 2024.

\bibitem[Carion et~al.(2020)Carion, Massa, Synnaeve, Usunier, Kirillov, and Zagoruyko]{carion2020end}
N.~Carion, F.~Massa, G.~Synnaeve, N.~Usunier, A.~Kirillov, and S.~Zagoruyko.
\newblock End-to-end object detection with transformers.
\newblock In \emph{European conference on computer vision}, pages 213--229. Springer, 2020.

\bibitem[Sohn et~al.(2015)Sohn, Lee, and Yan]{sohn2015learning}
K.~Sohn, H.~Lee, and X.~Yan.
\newblock Learning structured output representation using deep conditional generative models.
\newblock \emph{Advances in neural information processing systems}, 28, 2015.

\bibitem[Kingma et~al.(2013)Kingma, Welling, et~al.]{kingma2013auto}
D.~P. Kingma, M.~Welling, et~al.
\newblock Auto-encoding variational bayes, 2013.

\bibitem[Oquab et~al.(2023)Oquab, Darcet, Moutakanni, Vo, Szafraniec, Khalidov, Fernandez, Haziza, Massa, El-Nouby, et~al.]{oquab2023dinov2}
M.~Oquab, T.~Darcet, T.~Moutakanni, H.~Vo, M.~Szafraniec, V.~Khalidov, P.~Fernandez, D.~Haziza, F.~Massa, A.~El-Nouby, et~al.
\newblock Dinov2: Learning robust visual features without supervision.
\newblock \emph{arXiv preprint arXiv:2304.07193}, 2023.

\bibitem[Li et~al.(2024)Li, Ren, Yang, Zhao, Wu, Xu, Bai, and Zhao]{li2024virt}
Z.~Li, L.~Ren, J.~Yang, Y.~Zhao, X.~Wu, Z.~Xu, X.~Bai, and H.~Zhao.
\newblock Virt: Vision instructed transformer for robotic manipulation.
\newblock \emph{arXiv preprint arXiv:2410.07169}, 2024.

\bibitem[Borucka(2019)]{borucka2019city}
J.~Borucka.
\newblock City walk: a didactic innovative experiment in architectural education.
\newblock \emph{World Transactions on Engineering and Technology Education}, 17:\penalty0 158--163, 2019.

\bibitem[He et~al.(2022)He, Chen, Xie, Li, Doll{\'a}r, and Girshick]{he2022masked}
K.~He, X.~Chen, S.~Xie, Y.~Li, P.~Doll{\'a}r, and R.~Girshick.
\newblock Masked autoencoders are scalable vision learners.
\newblock In \emph{Proceedings of the IEEE/CVF conference on computer vision and pattern recognition}, pages 16000--16009, 2022.

\bibitem[Vip()]{Viperx}
Viperx 300 robot arm 6dof.
\newblock \emph{URL https://www.trossenrobotics.com/viperx-300-robot-arm-6dof.aspx}.

\bibitem[Yang et~al.(2022)Yang, Liu, Li, Li, , and Sun]{streaming}
J.~Yang, S.~Liu, Z.~Li, X.~Li, , and J.~Sun.
\newblock Real-time object detection for streaming perception.
\newblock In \emph{Proceedings of the IEEE/CVF conference on computer vision and pattern recognition}, pages 5385--5395, 2022.

\end{thebibliography}

\clearpage
\section*{Appendix A: Hyperparameters}

As shown in Tab.~\ref{table:Hyperparameters}, we use the same hyperparameters as the ACT implements in the stronger IL model experiment. In the real-scene experiments, we adjusted the chunking size to 32.
In addition, we use DINOv2-small as the feature extractor to process image input with a smaller resolution and conduct MAE augmentation for the input images.
We use a probability of 0.7 to randomly augment the input image. If data augmentation is used, we randomly mask 50\% of the area for each image with different random square patch sizes.

\begin{table}[h]
    \small
    \begin{center}\renewcommand{\arraystretch}{1.01}
        \resizebox{0.92\linewidth}{!}{
		\begin{tabular}{lccc}
        \toprule
            \textbf{Hyperparameter} & \textbf{ACT (Sim)} & \textbf{Stronger ACT (Sim)} & \textbf{Stronger ACT (Real)} \\
            \midrule
            \# encoder layers & 4 & 4 & 4\\
            \# decoder layers & 7 & 7 & 7 \\
            \# heads & 8 & 8 & 8 \\
            feedforward dimension & 3200 & 3200 & 3200\\
            hidden dimension  & 512 & 512 & 512\\
            chunk size   & 100 & 100 & \textbf{32} \\
            learning rate  & 1e-5  & 1e-5 & 1e-5\\
            batch size     & 8 & 8 & 8\\
            beta     & 10 & 10 & 10\\
            dropout  & 0.1 & 0.1 & 0.1 \\
            MAE augmentation  & No & \textbf{Yes} & \textbf{Yes} \\
            feature extractor  & ResNet-18 & \textbf{DINOv2-small} & \textbf{DINOv2-small} \\
            input resolution & 480$\times$640 & \textbf{336$\times$406} & \textbf{336$\times$406} \\
            \bottomrule
        \end{tabular}
     }
    \end{center}
    \caption{\textbf{Hyperparameters} of ACT and Stronger ACT.}
    \label{table:Hyperparameters}
\end{table}

\begin{figure*}[h]
    \centering
    \includegraphics[width=0.85\linewidth]{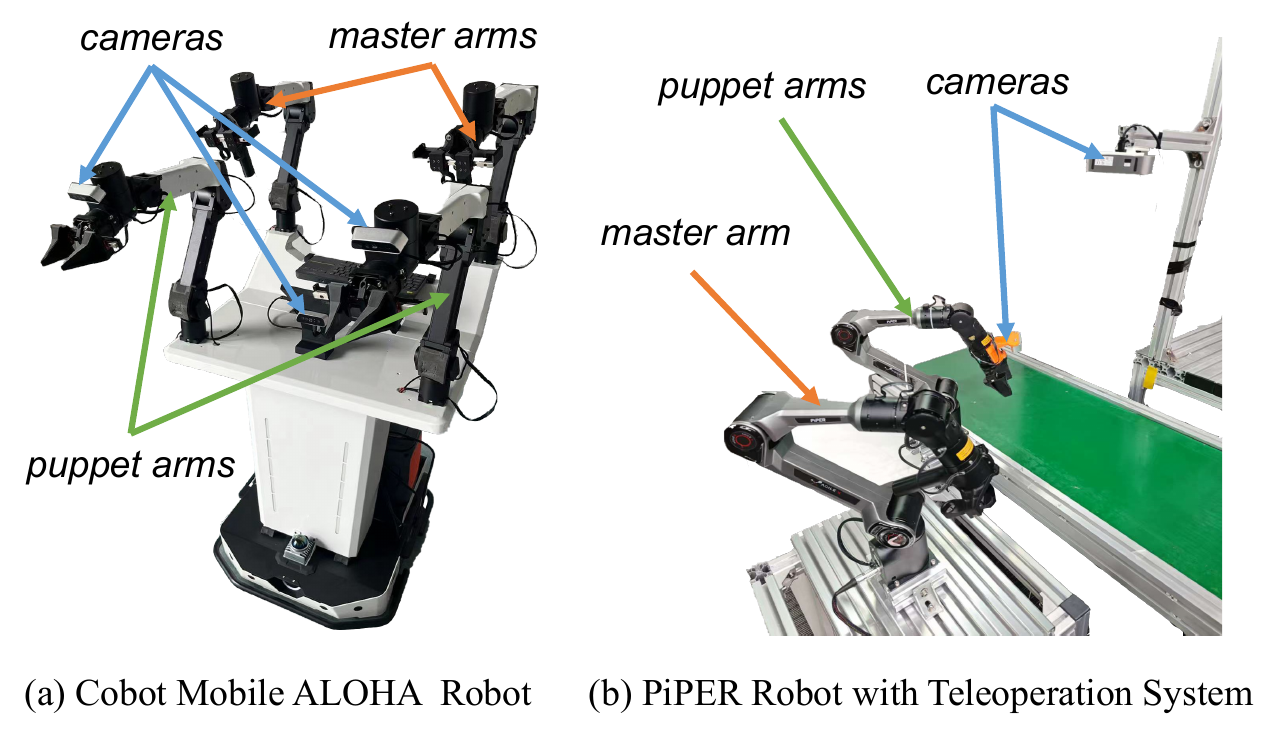}
    \vspace{-0.2cm}
    \caption{\textbf{Real-robot setups} including the Cobot mobile ALOHA robot and PiPER robot with teleoperation system.
 }\label{fig:real_robot}
\end{figure*}

\section*{Appendix B: Detailed Real-robot Setup}
As depicted in Fig.~\ref{fig:real_robot} (a), we use the Cobot mobile ALOHA robot for our experiments on real-world desktop tasks. It comprises two master arms, two puppet arms, and three RGB cameras. We move the master arm to a fixed pose and then use the camera fixed on it as the third-person view. Next, we control the left master arm and the left puppet arm for human demonstration, with the camera on the left puppet arm serving as the first-person view. As illustrated in Fig.~\ref{fig:real_robot} (b), we use the PiPER robot with a teleoperation system for our experiments on real-world unordered and mobile objects grasping tasks. It comprises one master arm, one puppet arm, one arm RGB camera, and one top-third-person camera. We collect all the data using a synchronous frame rate of 30hz, including camera images and the proprioceptions of the robot.

\section*{Appendix C: Video Demos}

All task results can be found in the \href{https://hd-space-robotics.github.io}{internet page} including \texttt{Grab teacup into box}, \texttt{Grab electronic pens placed randomly}, \texttt{Grab the bowl on belt with 8 cm/s moving speed}, and \texttt{Grab the bowl on belt with 16 cm/s moving speed}, \texttt{Grab spoon inside bowls on belt with 8 cm/s moving speed}, and \texttt{Grab spoon inside bowls on belt with 16 cm/s moving speed}.

\end{document}